\documentclass[twoside]{article}

\usepackage[accepted]{style}

\usepackage[round]{natbib}

\usepackage{enumerate}
\usepackage[utf8]{inputenc}
\usepackage[T1]{fontenc}
\usepackage{amsfonts}
\usepackage{amssymb}
\usepackage{amsthm}
\usepackage{amsmath}
\usepackage{colortbl}
\usepackage{dirtytalk}
\usepackage{mathtools}
\usepackage{subcaption}
\usepackage{tabularx}
\usepackage{multirow}
\usepackage{multicol}
\usepackage{tikz}
\usepackage{pgfplots}
\PassOptionsToPackage{hyphens}{url}
\usepackage[backref=false,colorlinks=true,final,linkcolor=black,urlcolor=red!60!black,
citecolor=cyan!50!black]{hyperref}
\usepackage[size=tiny,color=yellow]{todonotes}
\usepgfplotslibrary{statistics}
\pgfplotsset{compat=1.17}

\renewcommand{\vec}[1]{\ensuremath{\mathbf{#1}}}

\DeclareMathOperator*{\robust}{robust}
\DeclareMathOperator*{\rob}{rob}
\DeclareMathOperator*{\fair}{fair}

\DeclareMathOperator{\sign}{sgn}
\DeclareMathOperator{\HR}{HR}
\DeclareMathOperator{\OH}{OH}
\DeclareMathOperator{\CP}{CP}
\newcommand{\ud}{\triangleq}
\newcommand{\R}{\ensuremath{\mathbb{R}}}

\DeclareMathOperator{\RAF}{RAF}
\newcommand{\ra}{\rightarrow}

\newtheorem{theorem}{Theorem}[section]

\newtheorem{definition}[theorem]{Definition}
\newtheorem{corollary}[theorem]{Corollary}
\newtheorem{example}[theorem]{Example}
\newtheorem{remark}[theorem]{Remark}

\definecolor{color1}{HTML}{AD5042}
\definecolor{color2}{HTML}{3060AD}
\definecolor{color3}{HTML}{113061}
\definecolor{color4}{HTML}{89AD1F}
\definecolor{color5}{HTML}{4E6116}

\begin{document}

\twocolumn[

\styletitle{Abstract Interpretation-Based Feature Importance for SVMs}

\styleauthor{Abhinandan Pal \And  
    Francesco Ranzato\And  
    Caterina Urban\And  
    Marco Zanella}

\styleaddress{ IIIT Kalyani, India \And  University of Padova, Italy \And INRIA \& ENS $\mid$ PSL, France \And University of Padova, Italy} 
]

\begin{abstract}
We propose a symbolic representation for support vector machines (SVMs) by means of abstract interpretation, a well-known and successful technique for designing and implementing static program analyses. We leverage this abstraction in two ways: (1) to enhance the \emph{interpretability} of SVMs by deriving a novel feature importance measure, called \emph{abstract feature importance} (AFI), that does not depend in any way on a given dataset of the accuracy of the SVM and is very fast to compute, and (2) for verifying stability, notably \emph{individual fairness}, of SVMs and producing concrete counterexamples when the verification fails. We implemented our approach and we empirically demonstrated its effectiveness on SVMs based on linear and non-linear (polynomial and radial basis function) kernels. Our experimental results show that, independently of the accuracy of the SVM, our AFI measure correlates much more strongly with the stability of the SVM to feature perturbations than feature importance measures widely available in machine learning software such as permutation feature importance. It thus gives better insight into the trustworthiness of SVMs.
\end{abstract}

\section{Introduction}

Machine learning (ML) software is increasingly being employed in 
high stakes or sensitive applications \citep[etc.]{Chouldechova,khandani}. %
This poses important challenges for safety, privacy, and non-discrimination
\citep{buolamwini2018gender, obermeyer2019dissecting}. %
As a consequence, research in ML verification rapidly gained %
popularity \citep{liu2021algorithms, Survey} and demand for interpretable 
ML models is more and more pronounced \citep{XAI2021}. Notably, interpretability is needed to meet the requirements of recent legal regulations on sensitive automated decision-making applications, such as the General Data Protection Regulation (GDPR) and the Artificial Intelligence Act %
in the EU.

There is a tradeoff between %
interpretability and performance
of a ML model: %
non-linear models %
generally deliver much better predictions but they do not provide %
explanations. %
The most prominent interpretability technique is \emph{feature importance}, measuring %
the contribution of each input feature to a %
model prediction \citep{Bhatt2020}. %

\emph{Permutation feature importance} (PFI) \citep{breiman-random-forests,Fisher2019}, the most widely used and understood %
importance measure,
observes the decrease in predictive performance when a feature value is randomly shuffled: an increased loss is indicative of how much that feature is important for the predictive model. 
PFI is easy to explain, implement, and use, making it widely available in ML software (e.g., the scikit-learn Python library, %
etc.). On the other hand, %
the %
result may greatly vary depending on the %
dataset. %
Second, the result depends on shuffling and must be averaged across repetitions to stabilize, thus becoming resource intensive when the number of features is large.
Third, PFI yields misleading results when features are correlated \citep{Hooker2021}. Finally, and more importantly, the quality of the result heavily depends on the accuracy of the model! Notably, model variance to feature perturbations \citep{cacm18} and PFI are strongly correlated only when the model generalizes well.

\paragraph{Contributions.}
In this work, we propose a novel feature importance measure for SVMs, called \emph{abstract feature importance} (AFI), that: (a) does \emph{not} depend %
on a given dataset or the accuracy of the model, and (b) is extremely fast to compute, independently of the number of input features. We support both linear and non-linear kernels, in particular the polynomial and the radial basis function (RBF) kernels.

We derive our importance measure from a symbolic %
representation of a SVM based on \emph{abstract interpretation} \citep{CC77,cousot21}, a well-established framework for designing computable and correct %
over-approximations of model computations. Specifically, the concrete quantities being manipulated by model computations are represented using an \emph{abstract domain}, which defines their %
abstract counterparts and their data-structure representations, as well as algorithms to manipulate them according to the semantics of the computations. 

We leverage existing abstract domains such as hyperrectangles~\citep{CC77} and reduced affine forms \citep{RAF} that we combine with a novel abstract domain tailored for precisely representing computations with one-hot encoded categorical input features. %
We %
show the effectiveness of this combination %
in verifying model stability against feature perturbations. In particular, we focus on verifying \emph{individual fairness}~\citep{dwork2012fairness}.%
We evaluate our approach %
by verifying SVMs trained on the reference datasets in the literature on ML fairness~\citep{survey-fair} and considering different similarity relations.

Our approach is \emph{sound}, meaning that an individually fair abstract representation of a SVM %
implies that the %
SVM is also fair. Thus, the fraction of successful fairness verifications over a test dataset is a \emph{lower bound} on individual fairness of a SVM.
On the other hand, our approach is \emph{not complete} as there are cases in which the %
SVM is fair but the verification of its abstract representation fails due to imprecisions introduced by the abstraction. Our third contribution in this work is a way to leverage this abstract representation to generate concrete counterexamples when unable to verify fairness, i.e., concrete similar inputs to a SVM that result in different classifications. The fraction of successful counterexample searches over a test dataset yields a lower bound on how biased an SVM is and thus, by complement, an \emph{upper bound} on individual fairness of a SVM.

Finally, we conduct an extensive experimental comparison between our proposed importance measure AFI and the standard and popular PFI and we show that AFI is better correlated with stability of a SVM model to feature perturbations \emph{independently of the accuracy of the model}.

\paragraph{Related Work.}%

Feature importance measures can be \emph{local}, i.e., measuring feature importance for a specific prediction, or \emph{global}, i.e., measuring importance over the entire input space of the ML model. We also distinguish \emph{model-agnostic} measures, which can be applied to any model, and \emph{model-specific} measures. Finally, we
classify importance measures in \emph{performance-based}, i.e., measuring importance with respect to the predictive performance of the model (requiring knowledge of the ground truth values), 
and \emph{effect-based}, 
measuring importance based on the magnitude of change in the predicted outcome due to changes in the feature value
(requiring no knowledge of the ground truth values). %
PFI \citep{breiman-random-forests,Fisher2019} is a global, model-agnostic, performance-based measure. AFI, our novel feature importance measure, is \emph{specific for SVMs} but can be used \emph{both} as \emph{global and local} measure, and is \emph{effect-based}. %
Several other model-agnostic %
importance measures have been proposed in the literature. Prominent effect-based measures are %
visual tools such as partial dependence (PD) \citep{friedman2001greedy}, individual conditional expectation (ICE) \citep{Goldstein2015}, and accumulated local effects (ALE) \citep{Apley2020} plots. Other notable %
effect-based measures are Shapley values \citep{Strumbelj2014}, and local measures such as local interpretable model-agnostic explanations (LIME) \citep{Ribeiro2016}, and SHapley Additive exPlanations (SHAP) \citep{Lundberg2017}. 
Visual tools, such as individual conditional importance (ICI) and partial importance (PI) curves \citep{Casalicchio2018}, are also proposed for local %
performance-based measures. \cite{Casalicchio2018} additionally propose a Shapley feature importance (SFIMP) measure that allows comparing feature importances across different models.
Input gradient \citep{Hechtlinger2016} is a local %
measure that can be both effect-based and performance-based.
Feature importance measures specific for SVMs are typically limited to linear SVMs or face scalability issues with the number of features, e.g.\ \citep{mladenic2004feature,chang2008feature}. %
By contrast, our AFI measure also supports non-linear kernels and has no scalability issues. %

Our work generally contributes to the research ecosystem around the verification of ML models using formal methods \citep{liu2021algorithms,Survey}. 
Most approaches have focused on neural networks \citep[etc.]{roh20,ruoss2020learning,Urban20,yurochkin20} while here we focus on SVMs. %
Our work leverages the SVM verifier 
SAVer~\citep{ranzato2019robustness}. %
In addition, %
we introduce here a more precise abstraction for one-hot encoded features. %
Our fairness analysis is closely in line with the approach by \cite{ranzato2021fairness}, who evaluated the individual fairness of decision tree ensembles trained by a new fairness-aware learning technique. 
Similar works either consider a very different notion of fairness or a different ``threat model'', in most cases both. \cite{xiao2015support} evaluate security against flipping a few labels to maximize classification error. \cite{ghosh2022algorithmic} consider group and causal fairness metrics. \cite{langenberg2019robustness} deal with robustness of SVMs against adversarial attacks. \cite{fish2016confidence} propose a new fairness metric where they add a new feature with random values and bias individuals on this feature: the model is fair when it recovers the original labels. \cite{park2022privacy} put forward a protocol to protect sensitive information and train a fair model using homomorphic encryption. %
\cite{mehrabi2021survey} and \cite{verma2018fairness} discuss several fairness metrics used to verify a variety of ML models.

\section{Background}

\paragraph{Support Vector Machines.}\label{svm-sec}
SVMs \citep{cs00} are %
machine learning models based on %
separation curves that partition the input vector space into regions that best fit binary classification labels $L=\{-1,+1\}$. 
Separation curves are computed %
by maximizing their distance (margin) from the closest vectors in the training dataset. 
The simplest %
SVM is linear, which in its primal form boils down to an 
hyperplane $\vec{w} \cdot \vec{z} - b = 0$, 
where $\vec{w} \in \mathbb{R}^d$ and  $b \in \mathbb{R}$ are learned parameters, %
that
determines whether an input %
$\vec{z}$ falls above/below (i.e., $\sign(\vec{w} \cdot \vec{z} - b)=\pm 1$) 
w.r.t.\ the %
hyperplane. 
This approach is extended to non-linear SVMs through a  projection to a high-dimensional space 
via a kernel function 
$k: \mathbb{R}^d \times \mathbb{R}^d \rightarrow \mathbb{R}$.
Given a training dataset $T = \{(\vec{x}_1,y_1),....,(\vec{x}_n,y_n)\} \subseteq X \times \{-1,+1\}$, kernel function 
$k$ and learned parameters $c_i, b \in \mathbb{R}$, %
$i\in [1,n]$, a non-linear
 SVM %
 $C_T$ is represented in its dual form by the function
    $C_T(\vec{z}) \triangleq \sign \left( \sum^{n}_{i=1} \left( c_i y_i k(\vec{x}_i, \vec{z}) \right) - b \right)$.
Most common kernels are:
(i) \emph{linear}, where 
$k(\vec{x},\vec{z}) =\vec{x} \cdot \vec{z}$; 
(ii) \emph{polynomial}, where $k(\vec{x},\vec{z}) =(\vec{x} \cdot \vec{z} + c)^p$, for some hyperparameters $c\in \mathbb{R}$ and $p\in \mathbb{N}$; 
(iii) \emph{radial basis function} (RBF), where 
$k(\vec{x},\vec{z}) =e^{-\gamma\lVert \vec{x}-\vec{z}\rVert^2_2}$, for some
positive hyperparameter $\gamma>0$.
In multi-classification for a set of labels $L = \{y_1,...,y_m\}$, $m > 2$, 
the standard approach is a reduction into multiple binary classification problems %
combined by leveraging a voting 
procedure over %
different labels.  
\paragraph{Abstract Interpretation.}\label{ai-sec}
A tuple ${\langle A, \sqsubseteq^A, \gamma^A \rangle}$ is a \emph{numerical abstract domain} (or abstraction) when 
$\langle A, \sqsubseteq^A \rangle$ is an partially ordered set of abstract values 
and  ${\gamma^A: A \rightarrow \wp(\mathbb{R}^n)}$ is a concretization function which maps abstract values to sets of numerical vectors and monotonically preserves the ordering relation, i.e., 
${a_1 \sqsubseteq^A a_2}$ implies $\gamma^A(a_1) \subseteq \gamma^A(a_2)$. Intuitively, an 
abstract domain defines a symbolic representation of sets of vectors in $\wp(\mathbb{R}^n)$. 

Given a $k$-ary operation $f: (\mathbb{R}^d)^k \rightarrow \mathbb{R}^d$, for some $k >0$, 
a corresponding abstract function $f^A: A^k \rightarrow A$ is a \emph{sound} %
(over-)approximation of $f$ on $(a_1,...,a_k)\in A^k$ when 
$\{f(\vec{x}_1,...,\vec{x}_k) \mid \vec{x}_i \in \gamma^A(a_i) \} \!\subseteq\!\, \gamma^A(f^A (a_1,...,a_k))$
 holds. Morever, $f^A$ is a \emph{complete} (over-)approximation of $f$ on its input $(a_1,...,a_k)$
when equality holds. %

\paragraph{Abstract Domains.}
We consider the well-known abstract domain of \emph{hyperrectangles} (or \emph{intervals})~\citep{cousot21,rival-yi}. 
The hyperrectangle abstract domain $\HR_n$ consists of $n$-dimensional vectors $h$ of real intervals 
$h= \big([l_1, u_1], \ldots,[l_n, u_n]\big) \in \HR_n$,
with lower and upper bounds $l_i,u_i\in \mathbb{R} \cup \{-\infty,+\infty\}$ such that $l_i\leq u_i$. Hence, the concretization function $\gamma^{\HR}: \HR_n \rightarrow \wp(\mathbb{R}^n)$
is defined by $\gamma^{\HR}(h) \ud \{\vec{x}\in \mathbb{R}^n \mid \forall i.\, l_i \leq \vec{x}_i \leq u_i\}$.
Abstract operations are defined by extending the following abstract additions and multiplications of intervals:
$[l_1,u_1] +^{\HR} [l_2,u_2] \ud [l_1+l_2,u_1+u_2]$ and 
$[l_1,u_1] *^{\HR} [l_2,u_2] \ud [\min(l_1l_2, l_1u_2, l_2u_1, l_2u_2),\max(l_1l_2, l_1u_2, l_2u_1, l_2u_2)]$.

It is known that 
a compositional abstract evaluation on $\HR$ of an %
expression $\mathit{exp}$ 
can be imprecise, e.g.,
the %
evaluations of the simple expressions $x-x$ and $x\cdot x$ 
on an input interval $[-c,c]$, with $c >0$, yield, respectively, $[-2c,2c]$ and $[-c^2,c^2]$, rather than
the exact intervals $[0,0]$ and $[0,c^2]$. This dependency problem can trigger a significant source of imprecision 
for the hyperrectangle abstraction of a polynomial/RBF SVM classifier. 
Therefore, following \cite{ranzato2019robustness}, for our SVM abstract representations, we 
leverage the relational reduced affine form (RAF) abstraction.
A RAF for vectors in $\R^n$ is given 
by an expression 
$a_0 +\textstyle\sum_{i=1}^n a_i\epsilon_i + a_r \epsilon_r$,
where  the $\epsilon_i$'s 
are symbolic variables ranging in the real interval $[-1,1]$ representing a dependence from
the $i$-th component of the vector, while $\epsilon_r$ is a further 
symbolic variable in $[-1,1]$
which accumulates all the approximations introduced by non-linear operations. 
Thus, 
$\RAF_n  \ud \{ 
a_0 + \textstyle\sum_{i=1}^n a_i \epsilon_i + a_r \epsilon_a \mid %
a_0,a_1,...,a_n \in \R,\, a_r\in \R_{\geq 0}\}$. %
The concretization map $\gamma^{\RAF}: \RAF_n \rightarrow \wp(\mathbb{R})$ is defined by  
$\gamma^{\RAF}(a_0 + \textstyle\sum_{i=1}^n a_i \epsilon_i + a_r \epsilon_a) \ud %
\{ x\in \R \mid a_0-\sum_{i=1}^n |a_i|-|a_r| \leq x \leq a_0+\sum_{i=1}^n |a_i|+|a_r|\}$.
Moreover, $\RAF_n$ also has a top element $\top^{\RAF}$ representing the lack of information, i.e., such that 
$\gamma^{\RAF}(\top^{\RAF})=\mathbb{R}$.
Linear operations, namely additions and scalar multiplications, admit a complete approximation on the RAF abstraction. Thus, RAFs settle the dependency problem for linear expressions.
Instead, non-linear abstract operations, such as multiplication, must necessarily be approximated for RAFs. 
We will use an optimal abstract multiplication of RAFs defined  
by \cite{skalna2017}. 
\paragraph{Robustness.}
\label{sec:classifiers}
We consider an input space $X \subseteq \mathbb{R}^d$, a set of classification labels $L 
=\{y_1, ... ,y_m\}$, and a dataset $T = \{(\vec{x}_1,y_1),....,(\vec{x}_n,y_n)\} \subseteq X \times L$. A classifier trained on the  dataset $T$ is modeled as a map $C_T: X \rightarrow L$.

An adversarial region for an input sample $\vec{x}\in X$ is designated by a perturbation $P(\vec{x}) \subseteq X$ such that $\vec{x} \in P(\vec{x})$. Usually,  a perturbation function $P: X \rightarrow \wp(X)$ is defined through a metric $m$ to measure similarity between inputs as their distance w.r.t.\ $m$. 
The most common metric is induced by the $\ell_{\infty}$ maximum norm \citep{carlini} defined as $\lVert \vec{x}\rVert_{\infty} \triangleq \max\,\{\vec{x}_1,\ldots,\vec{x}_d\}$, so that 
the corresponding perturbation $P_{\infty}(\vec{x})$ includes all the vectors $\vec{z} \in X$ whose $\ell_{\infty}$  distance from 
$\vec{x}$ is bounded by 
a threshold $\epsilon \in \mathbb{R}^+$, that is, 
\(
P_{\infty}(\vec{x}) \triangleq \{ \vec{z} \in X \mid \lVert \vec{x}- \vec{z}\rVert_{\infty} \leq \epsilon  \}
\).

A classifier $C$ is robust (or stable) for a perturbation function $P$ on an input $\vec{x}$,
denoted by $\robust(C,\vec{x},P)$, 
when for all $\vec{z}\in P(\vec{x})$, $C(\vec{z}) = C(\vec{x})$ holds.
Robustness to a perturbation $P$ is used as a major metric  \citep{cacm18} to assess a classifier $C$ on a testing set 
$T \subseteq X \times L$ as follows:
    ${\textstyle \rob_T(C,P)} \triangleq \frac{|\{ (\vec{x},y) \in T~|~ \robust(C,\vec{x},P) \}|}{|T|}$.
\paragraph{SAVer.}\label{saver-sec}
Our work leverages SAVer (SVM Abstract Verifier), an automatic tool for robustness verification of SVMs introduced by \cite{ranzato2019robustness}. Given an SVM %
$C:X\ra L$, 
 SAVer leverages %
 an abstraction $A_n$ of $\wp(\mathbb{R}^n)$ 
to first achieve a sound %
abstraction $P^\sharp(\vec{x})\in A_n$ 
of an adversarial region $P(\vec{x})$, i.e., $P(\vec{x}) \subseteq \gamma^A(P^\sharp(\vec{x}))$, %
and then applies sound abstract versions of the transfer functions occurring in $C$ %
to design an abstract SVM $C^\sharp: A\ra \wp(L)$ 
that
computes an over-approximation of the labels assigned to inputs in $P(\vec{x})$, i.e.,
$\{C(\vec{z})\in L \mid \vec{z}\in P(\vec{x})\} \subseteq C^\sharp( P^\sharp(\vec{x}))$.
If  $C^\sharp( P^\sharp(\vec{x}))=\{y_i\}$, then %
every input in $P(\vec{x})$ is %
classified as $y_i$, so $C$ is proved robust over %
$P(\vec{x})$. 
In the binary %
case $L=\{-1,+1\}$, 
$C^\sharp$ consists of an abstract function $\mathcal{A}_C^\sharp:A_n \ra A_1$
that %
computes an over-approximation
$\mathcal{A}_C^\sharp(P^\sharp(\vec{x}))$ 
of the set of distances
between samples in $P^\sharp(\vec{x})$ and the %
separation curve, and then over-approximates %
the set of labels: %
$C^\sharp( P^\sharp(\vec{x})) \ud %
$ \textbf{ if } $\gamma^{A_1}(\mathcal{A}_C^\sharp(P^\sharp(\vec{x}))) \subseteq \mathbb{R}_{<0}$ \textbf{ then } $\{-1\}$ %
\textbf{ elseif } $\gamma^{A_1}(\mathcal{A}_C^\sharp(P^\sharp(\vec{x}))) \subseteq \mathbb{R}_{>0}$ \textbf{ then } $\{+1\}$ %
\textbf{ else } $\{-1,+1\}$.
In multi-classification, the voting %
also needs to be soundly approximated \citep{ranzato2019robustness}.
\section{Abstract Feature Importance}
\label{sec:FeatureInfluenceRanking}
We can now define our \emph{abstract feature importance} (AFI).

\begin{definition}[\textbf{Abstract Feature Importance}]\rm
Let $C: \mathbb{R}^n \rightarrow L$ be a SVM classifier and let $\mathcal{A}^{\RAF}_C:(\RAF_n)^n \ra \RAF_n$ be
its abstraction  in the RAF abstract domain. 
Let $\mathcal{A}^{\RAF}_C(f_1, \ldots, f_n) \ud
a_0 + \textstyle\sum_{i=1}^n a_i \epsilon_i + a_r\epsilon_r$ be the abstract computation output  
for an abstract input $(f_1, \ldots, f_n)$, $f_i \in \RAF_n$. 
The importance of every input feature $i\in [1,n]$ is defined as the absolute value $|a_i|\geq 0$. 
\qed
\end{definition}

The definition purposely approximates by ignoring the accumulative error due to the approximations of all non-linear operations performed by $C$, i.e., the term $a_r\epsilon_r$, influenced by all input features.
When $(f_1, \ldots, f_n)$, $f_i \in \RAF_n$, 
abstracts the whole input space $X \subseteq \mathbb{R}^n$, AFI measures the \emph{global} feature importance.
Otherwise, AFI measures the \emph{local} importance on the output label.

\begin{example}\rm 
Let us consider a toy linear SVM $C$ over a space $X \subseteq \mathbb{R}^2$ of values normalized to $[-1, +1]$, thus $X = \{\vec{x} \in \mathbb{R}^2 ~|~ -1 \leq x_1, x_2 \leq +1\}$. We consider two support vectors $\vec{v}_1 = (-0.5, 1), \vec{v}_2 = (0.5, -1) \in X$ labeled respectively as $-1, +1$ with weights $w_1 = w_2 = 0.5$ and bias $b = 0$, so that $C^{\text{sgn}}(\vec{x}) = -0.5 (\vec{v}_1 \cdot \vec{x}) + 0.5 (\vec{v}_2 \cdot \vec{x})$. We %
express $X$ as the RAF $a=(0 \pm \epsilon_1, 0 \pm \epsilon_2)\in (\RAF_2)^2$. %
By performing the abstract computations of $\mathcal{A}^{\RAF}_C$ on this input we obtain: $\mathcal{A}^{\RAF}_C(a) =\textstyle-0.5(\vec{v}_1\cdot^{\RAF} a) +0.5(\vec{v}_2\cdot^{\RAF} a) = \textstyle-0.5 (-0.5(0 \pm \epsilon_1) + 1(0 \pm \epsilon_2)) \textstyle+0.5 ( 0.5(0 \pm \epsilon_1) - 1(0 \pm \epsilon_2)) = \textstyle-0.5 (0 \pm (-0.5) \epsilon_1 \pm \epsilon_2) \textstyle+0.5 (0 \pm 0.5 \epsilon_1 \pm (-1) \epsilon_2) = 0 \pm 0.5 \epsilon_1 \pm (-1) \epsilon_2$.
We therefore infer the importance indices $|a_1| = 0.5$ and $|a_2| = 1$ for, respectively, $x_1$ and $x_2$, and conclude that $x_2$ is twice as important as $x_1$. Note that, since we considered a linear SVM, it can be rewritten in primal form as:
 $C^{\text{sgn}}(\vec{x})
 =
 \textstyle-0.5 (\vec{v_1} \cdot \vec{x}) + 0.5 (\vec{v_2} \cdot \vec{x}) %
 =
 \textstyle-0.5 (-0.5x_1 + x_2) + 0.5 (0.5x_1 -x_2) %
 =
 0.5x_1 - x_2 = (0.5, -1) \cdot \vec{x}$,
thus obtaining an explicit weight $\vec{w} = (0.5, -1)$ for the input features, whose absolute values $0.5, 1$ exactly match our importance indices.
\qed
\end{example}

Our importance indices depend on the size of the input, 
which can make results harder to read and interpret, especially when the number of features is high. %
We use a simple clustering strategy to assign them a score: we consider the distribution of feature importances and compute its mean $\mu$ and standard deviation $\sigma$, and we assign to each feature $x_i$ the score $score_i \in \mathbb{Z}$ such that $\mu + score_i \sigma \leq \frac{a_i - \mu}{\sigma} < \mu + (score_i + 1) \sigma$, which has the same effect of standardizing the distribution into a normal $\mathcal{N}(0, 1)$ and slicing the distribution at every unit, labeling every slice with a progressive number. By doing so, features moderately influencing the result will have a score close to zero, while relevant feature will have higher scores, and those not influencing the outcome will have a negative score. Last, we suggest to shift and clip such distribution in order to obtain grades, e.g., in $[3, 10]$, which can be achieved by a simple transformation: $grade_i = \max(\min(10, score_i + 6), 3)$. For instance, let us consider a distribution of indices $a_i$ given by $(1, 6, 2, 5, 6, 1, 6, 7, 8, 9)$, where $\mu = 5.1$ and $\sigma = 2.85$: we compute $gradei_i$ as $(5, 7, 5, 6, 7, 5, 7, 7, 8, 8)$. In this case it becomes easy to see that $x_1$ and $x_3$ have similar impact on the classification, although having different scores.

Last, we emphasize the fact that AFI does not require any knowledge on the ground truth values, nor the actual output of the classifier, as it focuses on the computation process performed by the classifier, rather than how the result of such computation is used to assign a label to a point, thus making this approach feasible in scenarios where the correct output is not known in advance.

\section{An Abstraction for One-Hot Encoding}
\label{sec:one-hot}
ML algorithms 
need 
a way to represent categorical data in numeric form. 
Let $F = \{c_1, c_2, \ldots, c_k\}$ be the set of values of a categorical feature $f$.
Assigning a %
number to each value in $F$ %
introduces an unwanted ordering relation among features. %
A better %
approach is \emph{one-hot encoding}, that is,
replacing 
$f$ with 
 $k$ binary %
features $(x_1^f, x_2^f, \ldots, x_k^f) \in \{0, 1\}^k$ such that $\forall i \in [1,k].\, 
x_i^f = 1 \Leftrightarrow f = c_i$. This sequence of bits is also referred to as a \emph{tier} of $f$.

Abstractions %
such as the hyperrectangle and RAF abstract domains, are likely to suffer from a significant loss of precision when dealing with one-hot encoded features, as they are not able to keep track of the relationship existing 
between the binary features resulting from the encoding. 

\begin{example}\rm 
 \label{ex:precision-loss}
Let us consider a categorical feature $f\in F =\{\mathit{red}, \mathit{green}, \mathit{blue}\}$ 
and let $(x_r,x_g,x_b)\in \{0,1\}^3$ be the corresponding one-hot encoded tiers. %
Consider the set $\{\mathit{red}, \mathit{green}\}$,
represented by the set of tiers $X = \{(1, 0, 0), (0, 1, 0)\}$. The most precise hyperrectangle abstraction of $X$ 
is $h = (x_r \in [0, 1], x_g\in [0, 1], x_b \in [0,0])\in \HR_3$. 
Observe that $h$ also represents infinitely many vectors in $\mathbb{R}^3$ that do not belong to $X$ and 
are illegal encodings, such as $(0.3, 0.7, 0)$, $(1, 1, 0)$ or $(0, 0, 0)$.
 \qed
\end{example}

To hinder this loss of precision, we define the \emph{One-Hot abstraction} $\OH$, a novel 
family of numerical abstractions tailored for one-hot encoded values.

Let us first recall the %
\emph{constant propagation} abstract domain $\CP\ud \mathbb{R} \cup \{\bot^{\CP},\top^{\CP}\}$, %
ordinarily used %
by modern compilers \citep{dragonbook,CP91}. $\CP$ is a flat domain 
whose partial order $\sqsubseteq^{\CP}$ 
is defined by $\bot^{\CP} \sqsubseteq^{\CP} z \sqsubseteq^{\CP} \top^{\CP}$, for all $z\in \mathbb{R}$. 
The concretization $\gamma^{\CP} : \CP \rightarrow \wp(\mathbb{R})$ is: $\gamma^{\CP}(z)\ud\{z\}$, 
for all $z\in \mathbb{R}$, meaning that a given numerical feature can only assume
a constant value $z$,
$\gamma^{\CP}(\top^{\CP})\ud \mathbb{R}$ %
representing no 
constancy information, 
while
$\gamma^{\CP}(\bot^{\CP})\ud\varnothing$ encodes unfeasibility. 
$\CP$ also has an abstraction map $\alpha^{\CP}:\wp(\mathbb{R}) \rightarrow \CP$ that 
provides the best approximation in $\CP$, i.e.\ least w.r.t.\ the order $\sqsubseteq^{\CP}$, 
of a set of values, which is: %
 $\alpha^{\CP}(\varnothing) \ud \bot^{\CP}$, $\alpha^{\CP}(\{z\}) \ud z$, and $\alpha^{\CP}(X) \ud \top^{\CP}$ otherwise.

The
One-Hot abstract domain for a $k$-dimensional one-hot encoded feature space is: %
$\OH_k \ud (\CP \times \CP)^k$.
Thus, abstract values are $k$-tuples of pairs of values in $\CP$, that %
keep track of the numerical information originated from a single one-hot $k$-encoded feature, both when this was originally false, i.e.\ $0$, 
or true, i.e.\ $1$. Given $a\in \OH_k$ and a component $i\in [1,k]$, let $a_{i, f/t}\in \CP$ 
denote, resp., the first/second element of the $i$-th pair in $a$.
The partial order $\sqsubseteq$ of $\OH_k$ is induced componentwise by $\sqsubseteq^{\CP}$, i.e., for all $a,b\in \OH_k$,
$a\sqsubseteq b \,\Leftrightarrow\, \forall i\in [1,k].\, a_{i, f} \sqsubseteq^{\CP} b_{i, f} \:\&\: 
a_{i, t} \sqsubseteq^{\CP} b_{i, t}$.
Then, for each component $i\in [1,k]$, the map $\hat{\gamma}_i: \OH_k \rightarrow \wp(\mathbb{R}^k)$ is defined as: 
$\hat{\gamma}_i(a) \!\ud\! 
\{ \vec{x} \in \mathbb{R}^k ~|~ \vec{x}_i \in \gamma^{\CP}(a_{i, t}),\, \forall j \neq i.\, 
\vec{x}_j \in \gamma^{\CP}(a_{j, f})\}$.
Thus, $a\in \OH_k$ represents through $\hat{\gamma}_i$ %
the set of tiers whose $i$-th component 
was originally set to true. Note that if, for some $i\in [1,k]$, either $a_{i,f}=\bot^{\CP}$ or
 $a_{i,t}=\bot^{\CP}$, then  $\hat{\gamma}_i(a)=\varnothing$. 
 A value $a\in \OH_k$ such that,
for all $i\in [1,k]$ and $u \in \{f, t\}$, $a_{i, u} \in\CP\smallsetminus \{\top^{\CP}\}$ is called \emph{top-less}.  
To retrieve all the  concrete vectors represented by $a$, 
we collect all the  vectors obtained by assuming that any component of the tier was originally set to true, namely, 
the concretization map 
$\gamma^{\OH_k}: \OH_k \rightarrow \wp(\mathbb{R}^k)$ is:
$\gamma^{\OH_k}(a) \ud \cup_{i = 1}^k \hat{\gamma}_i(a)$.

\begin{example}\rm 
 \label{ex:oh-no-precision-loss}
Let us continue Example~\ref{ex:precision-loss} by 
considering $a = \big((0, 1), (0, 1), (0, \bot^{\CP})\big) \in \OH_3$ as an abstraction of %
$X = \{(1, 0, 0), (0, 1, 0)\}$. %
Its concretization is: %
   $\gamma^{\OH_3}(a) %
   = \hat{\gamma}_1(a) \cup \hat{\gamma}_2(a) \cup \hat{\gamma}_3(a) %
  = \{\vec{x} \in \mathbb{R}^3 ~|~ \vec{x}_1 \in \{1\},\, \vec{x}_2 \in \{0\},\,  \vec{x}_3 \in \{0\}\} %
   \cup \{\vec{x} \in \mathbb{R}^3 ~|~ \vec{x}_1 \in \{0\},\, \vec{x}_2 \in \{1\},\,  \vec{x}_3 \in \{0\}\} %
   \cup \{\vec{x} \in \mathbb{R}^3 ~|~ \vec{x}_1 \in \{0\},\,  \vec{x}_2 \in \{0\},\,  \vec{x}_3 \in \varnothing\} %
  = \{(1, 0, 0), (0, 1, 0)\}$.
Thus $a$ precisely
 represents $X$. %
 \qed
\end{example}

Example~\ref{ex:oh-no-precision-loss} is not fortuitous. In fact,  
for any set $X$ of 
one-hot encoded tiers there always exists 
an abstract value $a$ in $\OH$ which precisely represents this set, i.e.,
$\gamma^{\OH}(a)=X$. 
\begin{theorem}
 \label{th:oh-alpha-prime}
If $X \in \wp(\mathbb{R}^k)$ is 
such that every vector of $X$ is a one-hot encoded tier $(0,...,0,1,0,...0)$, 
then the abstract value $a^X \in \OH_k$ defined as %
$a^X_i \ud  \big(%
\alpha^{\CP}(\{\vec{x}_i \mid \vec{x}\in X, \vec{x}_i = 0\}), %
\alpha^{\CP}(\{\vec{x}_i \mid \vec{x}\in X, \vec{x}_i = 1\})\big)$,
for all $i\in [1,k]$,
precisely represents $X$.
\end{theorem}

\begin{remark}\rm 
 \label{th:oh-alpha-no-top}
$a^X$ %
is always top-less 
because the components of each pair 
$a^X_i$ range in  $\{0,1,\bot^{\CP}\}$.
\qed
\end{remark}

Given a function $f: \mathbb{R} \rightarrow \mathbb{R}$, its  
abstract counterpart $f^{\CP}: \CP\rightarrow \CP$ on the CP domain is \citep{CP91}:
$f^{\CP}(z) \ud f(z)$, for all $z \in \mathbb{R}$,
$f^{\CP}(\bot^{\CP}) \ud \bot^{\CP}$, and $f^{\CP}(\top^{\CP}) \ud \top^{\CP}$.
In turn, %
$f^{\CP}$
allows us to define a sound abstract counterpart of $f$ on our $\OH$ abstraction:  
\begin{theorem}
 \label{th:oh-f-sound}
  A sound approximation  of $f$ on $\OH_k$ is  $f^{\OH}: \OH_k \rightarrow \OH_k$ defined, for all $i\in [1,k]$, as $(f^{\OH}(a))_i \ud \big(f^{\CP}(a_{i, f}), f^{\CP}(a_{i, t})\big)$.
\end{theorem}

\begin{example}\rm
 \label{ex:f-sound}
 Let us carry on Example~\ref{ex:oh-no-precision-loss}. 
 We apply %
 $f(x) \ud x^2 -3x +1$ to every %
 component %
 in $\gamma^{\OH_3}(a)$: $f(\gamma^{\OH_3}(a)) = \{(f(1), f(0), f(0)), (f(0), f(1), f(0))\} = \{(-1, 1, 1), (1, -1, 1)\}$. 
 By applying Theorem~\ref{th:oh-f-sound} to %
 $a$, %
we obtain
 $a' \ud f^{\OH}(a) = \big( (f^{\CP}(0), f^{\CP}(1)),$
 $(f^{\CP}(0), f^{\CP}(1)), (f^{\CP}(0), f^{\CP}(\bot^{\CP}) \big) =$ $\big((1, -1),$ $(1, -1),$ $(1, \bot^{\CP})\big)$,
 whose concretization is:
   $\gamma^{\OH_3}(a') %
   =$ $\hat{\gamma}_1(a') \cup \hat{\gamma}_2(a') \cup \hat{\gamma}_3(a') %
   = \{\vec{x} \in \mathbb{R}^3 ~|~ \vec{x}_1 \in \{-1\},\, \vec{x}_2 \in \{1\},\,  \vec{x}_3 \in \{1\}\} %
   \cup \{\vec{x} \in \mathbb{R}^3 ~|~ \vec{x}_1 \in \{1\},\, \vec{x}_2 \in \{-1\},\,  \vec{x}_3 \in \{1\}\} %
   \cup \{\vec{x} \in \mathbb{R}^3 ~|~ \vec{x}_1 \in \{1\},\,  \vec{x}_2 \in \{1\},\,  \vec{x}_3 \in \varnothing\} %
   = \{(-1, 1, 1), (1, -1, 1)\}$. 
Then, notice that soundness holds because $f(\gamma^{\OH_3}(a)) \subseteq \gamma^{\OH_3}(f^{\OH}(a))$. 
\qed 
\end{example}

Note that in Example~\ref{ex:f-sound}, $f(\gamma^{\OH_3}(a)) = \gamma^{\OH_3}(f^{\OH}(a))$ also holds, i.e., $f^{\OH}$ is a complete approximation of $f$ on $a$. This is as a consequence of the following general result.
\begin{corollary}\label{th:oh-f-complete}
 Let $a \in \OH_k$ be  top-less. Then (i) $f^{\OH}$ is a complete abstraction of $f$ on $a$; and (ii) given $f_1, f_2, \ldots, f_p: \mathbb{R} \rightarrow \mathbb{R}$, $f^{\OH}_1 \circ f^{\OH}_2 \circ \cdots \circ f^{\OH}_p$ is a complete abstraction of $f_1 \circ f_2 \circ \cdots \circ f_p$ on $a$.
\end{corollary}

We implemented this OH abstraction on top of (the interval and) RAF abstraction in SAVer.
 Given a categorical feature $f$ we first perform one-hot encoding, obtaining $(x_1^f, x_2^f, \ldots, x_k^f) \in \{0, 1\}^k$. When perturbing one-hot encoded values we allow every binary feature in the encoding to be either $0$ or $1$, so %
their abstract value is always in the shape $(0^{CP}, 1^{CP})^k$. As a consequence each abstract binary feature $a_i$ can be represented as the RAF $0.5 \pm 0.5 \epsilon_i$. 
We %
 keep track of the relation between features and tiers %
 using a global lookup table. After computing the abstract kernel, the resulting RAF $a$ contains information both from the regular and the OH analysis. 
 For verification purposes, we condense the latter 
 to a single interval representing the behavior of the categorical feature $f$: we compute $\gamma^{OH_k}$ over each tier
 and 
 we select the values minimizing and maximizing the RAF expression $a$, thus computing a sound approximation of the interval of variation induced by $f$. %
 Then, we build a RAF $a'$ by transferring numerical features as they are, and replacing tier information with the newly-computed intervals, expressed in RAF form. Finally, the resulting $a'$ allows to compute the superset of output labels as before.
Both the RAF $a$ or the condensed RAF $a'$ are suitable for our importance measure AFI: in the former case, unlike PFI, we are additionally able to measure the importance of tiers of a one-hot encoded feature.

\section{Individual Fairness}
\label{sec:individual-fairness}
Several formal models of fairness have been investigated in the literature.   
\cite{dwork2012fairness} 
point out several weaknesses of group fairness and therefore study 
\emph{individual fairness} defined as ``the principle that two individuals who are similar with respect to a particular task should be classified similarly'' \cite[Section~1.1]{dwork2012fairness}. This is  formalized as a
Lipschitz condition of the classifier, that is,  
by requiring that two individuals $\vec{x}, \vec{y} \in X$ whose distance is $\delta(\vec{x}, \vec{y})$, are mapped, 
respectively, to distributions $D_{\vec{x}}$ and $D_{\vec{y}}$ whose distance is at most  $\delta(\vec{x}, \vec{y})$. 
Intuitively, the output distributions for $\vec{x}$ and $\vec{y}$ are indistinguishable up to their distance.
Several distance metrics $\delta\colon X \times X \rightarrow \mathbb{R}_{\geq 0}$ can be used in this context, where 
\citep[Section~2]{dwork2012fairness}
studies the total variation or relative $\ell_\infty$ distances.

Following 
\cite{dwork2012fairness}, 
a classifier $C\colon X \rightarrow L$ is (individually) \emph{fair} when $C$ outputs the same label
for all pairs of individuals $\vec{x}, \vec{y} \in X$ satisfying a similarity relation $S \subseteq X \times X$ between input samples. This relation $S$ can be derived from a distance $\delta$ as follows: 
$(\vec{x}, \vec{y}) \in S \Leftrightarrow \delta(\vec{x}, \vec{y}) \leq \epsilon$, where $\epsilon \in \mathbb{R}$ is a similarity threshold. 

\begin{definition}[\textbf{Individual Fairness}]\rm
 \label{def:individual-fairness}
 A classifier $C\colon X \rightarrow L$ is \emph{fair} on an individual $\vec{x} \in X$ with respect to a similarity relation $S \subseteq X\times X$, denoted by $\fair(C, \vec{x}, S)$, when 
  $\forall \vec{z} \in X.\, (\vec{x}, \vec{z}) \in S \Rightarrow C(\vec{z}) = C(\vec{x})$. \qed
\end{definition}

To define a fairness metric for a classifier $C$, 
we %
compute how often $C$ is fair on sets of similar individuals in a test set 
$T \subseteq X \times L$:
 ${\textstyle\fair_{T,S}(C)} \ud \frac{|\{(\vec{x}, y) \in T ~|~ \fair(C, \vec{x}, S)\}|}{|T|}$.
Hence, individual fairness for a similarity relation $S$ boils down to robustness on the perturbation $P_S(\vec{x}) \ud \{\vec{z} \in X ~|~ (\vec{x}, \vec{z}) \in S\}$ induced by $S$, i.e., for all $\vec{x} \in X$, $\fair(C, \vec{x}, S) \Leftrightarrow\ \robust(C, \vec{x}, P_S)$.

\section{Mitigating Incompleteness}\label{sec:cex}
The abstract framework described in Sec. \ref{saver-sec} and \ref{sec:one-hot} is \emph{sound}, thus a classifier $C$ verified as \emph{robust} over a region $P(\vec{x})$ guarantees that every point $\vec{x'} \in P(\vec{x})$ receives the same label. The converse is generally not true for non-linear kernels, %
due to lack of \emph{completeness}: when the abstract verification is not able to assert robustness, it may be either due to a loss of precision or an actual point in $P(\vec{x})$ which receives a different label. We refer to the latter as a \emph{counterexample}. %
In case of an inconclusive analysis we can mitigate the effect of incompleteness by searching for counterexamples: if at least one is found the classifier can be marked as \emph{not robust}.
Finding a counterexample within a possibly infinite set of points however is a daunting task. %
Let $a \in \RAF_n$ be a sound abstraction for $P(\vec{x})$, $C$ a classifier, and $\mathcal{A}_C^{\RAF}$ its abstraction.
We define an informed heuristic search approach which leverages our AFI measure: \\ %
 1. let $a_{out} = a_0 + \textstyle\sum_{i=1}^n a_i \epsilon_i + a_r\epsilon_r$ be the output of the abstract computation on $\RAF$ (cf. Sec.~\ref{sec:FeatureInfluenceRanking}); \\
 2. if $C(\vec{x}) < 0$, we look for a potential counterexample $\vec{x^*}$ by maximizing $a_{out}$, i.e., selecting the maximum possible value for every $x_i$ when $a_i > 0$, and the minimum when $a_i < 0$ (the converse if $C(\vec{x}) > 0$); \\
 3. if $C(\vec{x^*}) \neq C(\vec{x})$, then $\vec{x^*}$ is a counterexample for $\vec{x}$, and the classifier is not robust; \\
 4.~otherwise we select the most influential feature $x_M$, and its mean value in $P(\vec{x})$ given by $m = \frac{\min\{x_M ~|~ \vec{x} \in P(\vec{x})\} + \max\{x_M ~|~ \vec{x} \in P(\vec{x})\}}{2}$, and we partition $P(\vec{x})$ using the cutting hyperplane $x_M \leq m$, obtaining left and right sets $P_{l}(\vec{x}), P_{r}(\vec{x}) \subseteq P(\vec{x})$; \\
 5.~we consider $a_l, a_r \in \RAF$ abstracting $P_l(\vec{x}), P_r(\vec{x})$, respectively, and we recursively repeat from step 1 until a counterexample is found, or a user-defined timeout is met.

Maximization in step 2 requires
additional care for features obtained through one-hot encoding, as exactly one must be set to $1$: %
we set to $1$ the most influential feature only. We also observe that computing $a_l, a_r$ during step 5 %
does not introduce any loss of precision, as $\RAF$ represent hyperrectangles, and partitioning one using a cutting hyperplane of the form $x_i \leq k$ yields two smaller hyperrectangles.

Step 1 and 2 correspond to looking for a counterexample in the vertices of the hyperrectangle represented by $a$, which have %
the greatest distance from the center and are therefore intuitively more likely to exhibit different labels. Since a hyperrectangle has $2^n$ vertices it is not feasible to check them all, and we thus use our feature importance analysis to infer a gradient pointing towards the most promising one. %
If no counterexample is found, it may be due to the separation curve of $C$ crossing $P(\vec{x})$ while leaving all the vertices on the same side. We therefore proceed to step 4 and 5 partitioning $P(\vec{x})$ into two smaller components $P_l(\vec{x})$ and $P_r(\vec{x})$ by cutting the former space in half along the axis of the most influential feature, and recursively repeating the process on the two components. Should a counterexample be found, $C$ can be definitively marked as not robust. Otherwise, we set a timeout mechanism, such as a limit on the recursion depth, to avoid non-termination. %
\begin{example}\rm
\label{ex:cex1}
 Let $\vec{s} = (0, -\sqrt{2}), \vec{t} = (-1, 1), \vec{v} = (1, 1)$ be the support vectors of an SVM with polynomial kernel $k(\vec{x}, \vec{y}) = (\vec{x} \cdot \vec{y} + 1)^2$, $-\alpha_\vec{s} = \alpha_\vec{t} = \alpha_\vec{v} = 1$ their weights, and $b = 0$ the bias. Classifier $C$ is therefore given by $C(\vec{x}) = -(\vec{s} \cdot \vec{x} + 1)^2 + (\vec{t} \cdot \vec{x} + 1)^2 + (\vec{v} \cdot \vec{x} + 1)^2$ and can be rewritten to its primal form $2 x_1^2 + 2(2 + \sqrt{2})x_2 + 1$, allowing to see the separation plane as the parabola $\Gamma: x_2 = -\frac{1}{2 + \sqrt{2}}x_1^2 - \frac{1}{2(2 + \sqrt{2})}$. We now consider $\vec{x'} = (0.5, -0.5)$ and $P(\vec{x'})$ the hyperrectangle of radius $0.5$ centered in $\vec{x'}$: $P(\vec{x}) = \{\vec{x} \in \mathbb{R}^2 ~|~ 0 \leq x_1 \leq 1, -1 \leq x_2 \leq 0\}$. We observe that $C(\vec{x'}) \approx -1.91$, but every point $\vec{x} \in P(\vec{x'})$ having $x_2 = 0$ will evaluate to the positive expression $2 x_1^2 + 1$, hence $C$ is not robust over $P(\vec{x})$. Visually, $\Gamma$ crosses $P(\vec{x})$ leaving some vertices on different sides of the space.

We consider $a = (0.5 \pm 0.5\epsilon_1, -0.5 \pm 0.5\epsilon_2) \in \RAF^2_2$ abstracting $P(\vec{x'})$, and compute $a_{out} = c \pm 0.5\epsilon_1 \pm (1+\sqrt{2})\epsilon_2 \pm d\epsilon_r$, where the values of the center $c \in \mathbb{R}$ and the non linear accumulation term $d \in \mathbb{R}$ are omitted as not relevant for this example. As a result we obtain the feature importance gradient $(0.5, 1+\sqrt{2})$ having both components positive. Since $C(\vec{x'}) < 0$ we are looking for positive counterexamples and, following step 3, we build $\vec{x^*} \in P(\vec{x'})$ by selecting the maximum values for $x_1, x_2$ in $P(\vec{x'})$, which yields $\vec{x^*} = (1, 0)$. Then compute $C(\vec{x^*}) = 2(1)^2 + 2(2 + \sqrt(2)(0) + 1  = 3 > 0$, hence a counterexample is found and $C$ can be marked as not robust.
 \qed
\end{example}

\begin{figure}[htb]
\centering
 \includegraphics[width=0.3\textwidth]{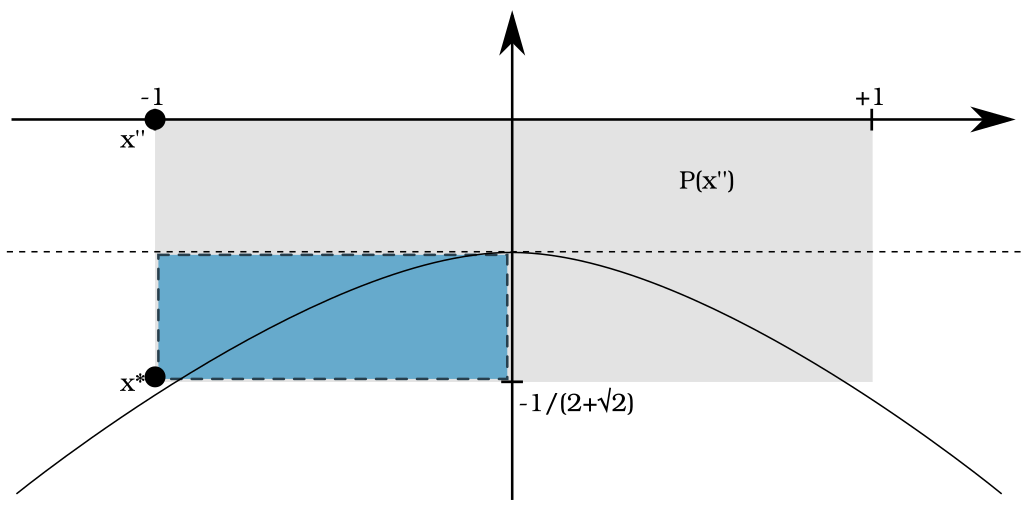}
 \caption{Recursion while searching for counterexamples.}
 \label{fig:counterexample}
\end{figure}

\begin{example}\rm
Continuing Example~\ref{ex:cex1}, we consider $\vec{x''} = (-1, 0) \in \mathbb{R}^2$, and the region $P(\vec{x''}) = \{\vec{x} \in \mathbb{R}^2 ~|~ -1 \leq x_1 \leq +1, -\frac{1}{2 + \sqrt{2}} \leq x_2 \leq 0\}$, as depicted by Fig. \ref{fig:counterexample}. It is easy to observe that every vertex of $P(\vec{x''})$ is on the same side of $\Gamma$ and thus receives the same label, while a small region in the bottom lays on the other side making $C$ not robust. We repeat the counterexample search process %
obtaining the importance gradient $(0, \frac{3 + 2\sqrt{2}}{(2+\sqrt{2})^2})$, this time looking for a negative counterexample hence moving in the opposite direction with respect to the gradient. By doing so we compute $\vec{x^*} = (-1, -\frac{1}{2 + \sqrt{2}})$ as the bottom-left corner, but $C(\vec{x^*}) = 1$ which does not produce a counterexample. We therefore follow steps 4 and 5 partitioning $P(\vec{x''})$ with $x_2 < -\frac{1}{2(2 + \sqrt{2})}$ as shown in the picture, and starting the recursion.
 After the first recursive step, none of the rectangles have counterexamples in their vertices, so another recursive call is started on the lowest one after partitioning on $x_1 < 0$. Both the resulting rectangles have now counterexamples in their vertices, which can be found by step 1 and 2. $C$ can thus be marked as non robust.
 \qed
\end{example}

\section{Experimental Evaluation}
\label{sec:experimental-evaluation}
We consider datasets standard to the fairness literature~\citep{survey-fair} (a) \textbf{Adult}~\citep{dua2017uci},  
which labels yearly incomes (above or below 50K US\$) based on personal attributes. (b) \textbf{Compas}~\citep{angwin2016machine}, 
which labels recidivism risk based on personal attributes and criminal history. (c) \textbf{Crime}~\citep{dua2017uci}, 
which labels communities below or above the median per capita violent crime rate based on socio-economic, law enforcement, and crime data. (d) \textbf{German}~\citep{dua2017uci}, which labels (good or bad) credit scores \citep{dua2017uci}. (e) \textbf{Health}~(\url{https://www.kaggle.com/c/hhp}), which labels ten-year mortality (above or below the median Charlson index) for a patient based on physician records and insurance claims.

The data is preprocessed following~\cite[Section~5]{ruoss2020learning}.
Some of these datasets exhibit a highly unbalanced label distribution, leading to high accuracy and $100\%$ individual fairness for a constant classifier like $C(\vec{x}) = 1$. Thus, following \citep{ruoss2020learning} we also report the \emph{balanced accuracy}, i.e.,
$\frac{1}{2} \left( \frac{\textit{truePos}}{\textit{truePos} + \textit{falseNeg}} + \frac{\textit{trueNeg}}{\textit{trueNeg} + \textit{falsePos}} \right)$.
\paragraph{Similarity Relations.}
\label{subsec:similarity}
Let $I \subseteq \mathbb{N}$ denote the set of features after one hot encoding and $\vec{x}, \vec{y} \in X$ be two individuals. Following \cite[Section~5.1]{ruoss2020learning}, we consider three  similarity relations.
\begin{description}
 \item [\textbf{\textsc{noise:}}] $S_{\textit{noise}}(x,y)$ iff $| {\vec{x}}_i - {\vec{y}}_i | \leq \epsilon$ for all $i \in I'$, and ${\vec{x}}_i = {\vec{y}}_i$ for all $i\in I\smallsetminus I'$, where $I' \subseteq I$ is a subset of numerical features. For our experiments, we consider $\epsilon = 0.05$ which leads upto a 10$\%$ perturbation for data normalised to [0,1].

 \item [\textbf{\textsc{cat:}}] $S_{\textit{cat}}(x,y)$ iff ${\vec{x}}_i = {\vec{y}}_i$ for all $i \in I\smallsetminus I'$, where $I' \subseteq I$ represent sensitive categorical attributes. For Adult and German, we select the gender attribute. For Compas, its race. For Crime, we consider state. Lastly, for Health, we consider gender and age group.

 \item [\textbf{\textsc{noise-cat:}}] $S_{\textit{noise-cat}}(x,y) = S_{\textit{noise}} \cup S_{\textit{cat}}(x,y)$ %
\end{description}

Further domain-specific similarities can be defined and handled by our approach by simply instantiating the underlying static analyzer %
with a suitable %
abstract domain. %
\paragraph{Setup.}

We trained the SVMs used in our experiments with scikit-learn~\citep{scikit-learn}. %
Hyperparameters were chosen by hit-and-trial and observing trends. 
\begin{figure}%
\begin{subfigure}{4cm}
  \begin{tikzpicture}
    \begin{axis}[height=3cm,
      width=3.5cm,
      grid=major,
      grid style={dotted},
      xlabel = degree,
      ylabel = coeff0,
      zlabel = precentage,
      legend style={at={(0.9,0.7)},anchor=south west,fill=none},
      mesh/cols = 5]
      \addplot3[surf, colormap/hot,opacity=0.4] file {./Plot-crime/BA-poly-reg1.txt};%
      \addplot3[surf, colormap/cool,opacity=0.4] file {./Plot-crime/RO-poly-reg1.txt};%
      \node at (rel axis cs:0.7,0,0.2) [above] {\footnotesize Fairness};
      \node at (rel axis cs:0.5,0,0.66) [above, rotate=-3] {\footnotesize Bal. Acc.};
    \end{axis}
  \end{tikzpicture}
\end{subfigure}
\begin{subfigure}{4cm}
  \begin{tikzpicture}
    \begin{axis}[height=3cm,
      width=3.5cm,
      grid=major,
      grid style={dotted},
      xlabel = reg,
      ylabel = gamma,
      y label style={at={(1,-0.3)}},
      zlabel = precentage,
      mesh/cols = 5]
      \addplot3[surf, colormap/hot,opacity=0.4] file {./Plot-crime/BA-rbf.txt};%
      \addplot3[surf, colormap/cool,opacity=0.7] file {./Plot-crime/RO-rbf.txt};%
      \node at (rel axis cs:0.85,0,0.8) [above , rotate=-27] {\footnotesize Fairness};
      \node at (rel axis cs:0.55,0,0.1) [above, rotate=-6] {\footnotesize Bal. Acc.};
    \end{axis}
    \end{tikzpicture}
\end{subfigure}
\caption{Trends on hyperparameters for SVMs with polynomial (left) and RBF (right) kernels trained on Crime.}
\label{fig:3Dplot}
\end{figure}
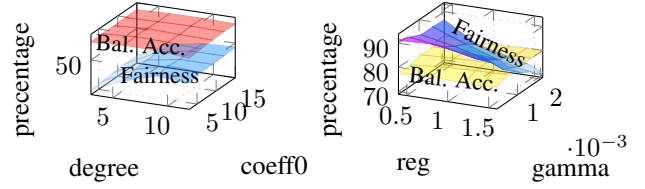
For instance, Fig. \ref{fig:3Dplot} shows trends in balanced accuracy and individual fairness with different hyperparameters for SVMs with polynomial and RBF kernels trained on the Crime dataset. Individual fairness is calculated with respect to the \textsc{noise-cat} similarity relation using the RAF+$\OH$ abstract domain. The plots show that balanced accuracy and individual fairness are inversely correlated and the change in fairness is steeper than accuracy. Similar trends occur for SVMs trained on the other datasets.
The final chosen hyperparameters for each kernel for each dataset were those that lead to SVMs with high balanced accuracy. 
Our implementation, the datasets with preprocessing pipelines, our SVM models and experiment scripts will be made available on GitHub upon publication of this work.

\paragraph{Individual Fairness.}

\begin{figure}[t]
 \centering
 \begin{tikzpicture}
  \begin{axis} [
   legend pos=south east,
   height=5.5cm, width=5.5cm,
   xmin=0, xmax=100,
   ytick={1,2,3,4},
   ytick style={draw=none},
   yticklabels={\footnotesize RBF, \footnotesize Polynomial, \footnotesize Linear},
   ymin=0.3,ymax=3.7,
   /pgfplots/boxplot/box extend=0.1
  ]
  \addplot+ [color1, thick, solid, fill=color1!20,      %
   boxplot prepared={
    median=79.838142,
    upper quartile=80.9179275,
    lower quartile=78.56703,
    upper whisker=80.981703,
    lower whisker=78.439479,
    draw position=2.55
   },
  ] coordinates {};
    \addplot+ [color2, thick, solid, fill=color2!20,     %
    boxplot prepared={
      median = 69.3,
      upper quartile = 89.22,
      lower quartile = 16.47,
      upper whisker = 90.72,
      lower whisker = 4.01,
      draw position = 3.3
    },
    ] coordinates {};
    \addplot+ [color3, thick, solid, fill=color3!20,    %
    boxplot prepared={
      median = 86.97,
      upper quartile = 92.3,
      lower quartile = 63.97,
      upper whisker = 92.73,
      lower whisker = 57.64,
      draw position = 3.1
    },
     ] coordinates {};
     \addplot+ [color4, thick, solid, fill=color4!20,      %
    boxplot prepared={
      median = 69.3,
      upper quartile = 89.22,
      lower quartile = 16.47,
      upper whisker = 90.72,
      lower whisker = 4.01,
      draw position = 2.9
    },
    ] coordinates {};
    \addplot+ [color5, thick, solid, fill=color5!20,       %
    boxplot prepared={
      median = 91.01,
      median = 86.97,
      upper quartile = 92.3,
      lower quartile = 63.97,
      upper whisker = 92.73,
      lower whisker = 57.64,
      draw position = 2.7
    },
     ] coordinates {};     
     
    \addplot+ [color1, thick, solid, fill=color1!20,  %
    boxplot prepared={
      median=77.64,
      upper quartile=78.19,
      lower quartile=74.31,
      upper whisker=79.46,
      lower whisker=73.53,
      draw position=1.55
    },
    ] coordinates {}; 
      \addplot+ [color2, thick, solid, fill=color2!20,     %
    boxplot prepared={
      median = 5.01,
      upper quartile = 14.97,
      lower quartile = 0.37,
      upper whisker = 21.80,
      lower whisker = 0.0,
      draw position = 2.3
    },
    ] coordinates {};
    \addplot+ [color3, thick, solid, fill=color3!20,    %
    boxplot prepared={
      median = 5.01,
      upper quartile = 14.97,
      lower quartile = 0.37,
      upper whisker = 21.80,
      lower whisker = 0.0,
      draw position = 2.1
    },
     ] coordinates {};
     \addplot+ [color4, thick, solid, fill=color4!20,      %
    boxplot prepared={
      median = 30.70,
      upper quartile = 36.78,
      lower quartile = 9.15,
      upper whisker = 37.85,
      lower whisker = 3.51,
      draw position = 1.9
    },
    ] coordinates {};
    \addplot+ [color5, thick, solid, fill=color5!20,       %
    boxplot prepared={
      median = 36.34,
      upper quartile = 42.36,
      lower quartile = 15.30,
      upper whisker = 44.36,
      lower whisker = 7.77,
      draw position = 1.7
    },
     ] coordinates {};

     \addplot+ [color1, thick, solid, fill=color1!20,     %
    boxplot prepared={
      median=78.19,
      upper quartile=79.74,
      lower quartile=75.51,
      upper whisker=80.73,
      lower whisker=65.71,
      draw position=0.55
    },
    ] coordinates {};
      \addplot+ [color2, thick, solid, fill=color2!20,     %
    boxplot prepared={
      median = 0.25,
      upper quartile = 16.29,
      lower quartile = 0.0,
      upper whisker = 48.37,
      lower whisker = 0.0,
      draw position = 1.3
    },
    ] coordinates {};
    \addplot+ [color3, thick, solid, fill=color3!20,    %
    boxplot prepared={
      median = 1.5,
      upper quartile = 33.08,
      lower quartile = 0.0,
      upper whisker = 68.67,
      lower whisker = 0.0,
      draw position = 1.1
    },
     ] coordinates {};
     \addplot+ [color4, thick, solid, fill=color4!20,      %
    boxplot prepared={
      median = 1.5,
      upper quartile = 6.51,
      lower quartile = 0.25,
      upper whisker = 54.89,
      lower whisker = 0.0,
      draw position = 0.9
    },
    ] coordinates {};
    \addplot+ [color5, thick, solid, fill=color5!20,       %
    boxplot prepared={
      median = 81.70,
      upper quartile = 89.47 ,
      lower quartile = 61.15,
      upper whisker = 92.98,
      lower whisker = 4.0,
      draw position = 0.7
    },
     ] coordinates {};
  \end{axis}
  \matrix [right] at (current bounding box.east) {
    \node [shape=rectangle, draw=color2, fill=color2!20, line width=1,label=right:\footnotesize Interval] {}; \\
    \node [shape=rectangle, draw=color3, fill=color3!20, line width=1,label=right:\footnotesize Interval+OH] {}; \\
    \node [shape=rectangle, draw=color4, fill=color4!20, line width=1,label=right:\footnotesize RAF] {}; \\
    \node [shape=rectangle, draw=color5, fill=color5!20, line width=1,label=right:\footnotesize RAF+OH] {}; \\
 \node [shape=rectangle, draw=color1, fill=color1!20, line width=1,label=right:\footnotesize Bal. Acc.] {}; \\
  };
 \end{tikzpicture}
 \caption{Comparison of individual fairness using different abstract domains for SVMs trained on Crime.}
 \label{box:Crime}
\end{figure}
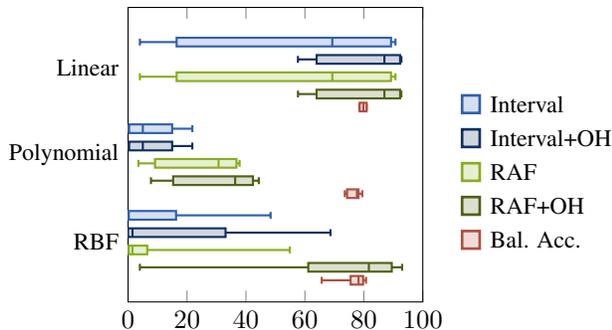

We show a summary %
of individual fairness and balanced accuracy scores obtained for SVMs trained on the Crime dataset in Fig.~\ref{box:Crime}. We can see that the RAF abstraction typically outperforms intervals, and our $\OH$ abstraction always yields equal or higher individual fairness score. 
The full raw data for each dataset shows the same trend. It is shown in Table~\ref{Tab:OH} in Appendix~\ref{appendix-B}.

\begin{table}[t]
\begin{center}
\caption{Bounds on fairness with respect to the \textsc{NOISE-CAT} similarity relation using the RAF+OH abstraction.}\label{tab:bounds}
\resizebox{1\columnwidth}{!}{
\begin{tabular}{ |p{1.2cm}|p{0.7cm}|p{0.7cm}|p{0.7cm}|p{0.7cm}|p{0.7cm}|p{0.7cm}|  }
 \hline & \multicolumn{2}{|c|}{\textbf{Linear}} & \multicolumn{2}{|c|}{\textbf{Polynomial}} & \multicolumn{2}{|c|}{\textbf{RBF}}\\
 \cline{2-7}
 \textbf{Dataset} & LB & UB & LB & UB & LB &UB \\
 \hline \hline
 Crime & 29.6 & 58.4  & 39.1 & 72.9 & 91.0 & 92.5 \\
 \hline
 Health & 95.5 & 99.6  & 0.02 & 98.7 & 29.4 & 97.4  \\
 \hline
 Compas & 94.9 & 94.9  & 0.09 & 71.4 & 89.3 & 93.0   \\
 \hline
 German & 81.5 & 81.5  & 10.0& 76.0& 0.0& 84.0   \\
 \hline
 Adult & 91.6 & 91.6  & 0.03& 89.5& 92.2 & 95.4   \\
 \hline
\end{tabular}
}\ \ \
\end{center}
\end{table}

In Table~\ref{tab:bounds}, for each dataset, we show bounds on individual fairness with respect to the  \textsc{NOISE-CAT} similarity relation using the RAF+OH abstraction: the lower bound is the verified individual fairness score (cf. Section~\ref{sec:individual-fairness}) and the upper bound is the estimate obtained with our counterexample search without input partitioning (cf. Section~\ref{sec:cex}), e.g., an upper bound of $76\%$ indicates that we found concrete counterexamples to individual fairness for $48$ over $200$ test data points. The gap between bounds is quite narrow for linear SVMs as well as for SVMs with RBF kernels trained on the Crime, Compas, and Adult datasets: in these cases, our RAF+OH abstraction is precise and our counterexample search heuristic is strong. On the other hand, the gap is much wider in the other cases, notably for SVMs with polynomial kernels, mostly due to a lower precision of the abstraction. Using partitioning up to $3.125\%$ of the original input size, yields similar upper bound values, which indicates the presence of few additional counterexamples. Only partitioning up to $0.1\%$ of the original size input size, we could find substantially more counterexamples.

\paragraph{Feature Importance.}

\begin{table}[t]
\caption{Comparison of our AFI and PFI on Compas.}\label{tbl:FI}
\begin{center}
\resizebox{1\columnwidth}{!}{

\begin{tabular}{ |p{1.65cm}|p{0.5cm}|p{0.5cm}|p{0.5cm}|p{0.5cm}|p{0.5cm}|p{0.5cm}|p{0.5cm}|p{0.5cm}|}
 \hline
\multirow{1}{4pt}{\textbf{Linear}} & LB & 87.2 & 93.3 & 97 & 97.1 & 97.7 & 99 & 100 \\ \cline{2-9}
\multicolumn{2}{|r|}{AFI (0.225s)} & 1 & 2 & 3 & 4 & 5 & 6 & 7 \\  
\multicolumn{2}{|r|}{PFI (1976s)} & 1 & 2 & 5 & 4 & 3 & 7 & 6 \\
 \hline
 \multirow{1}{4pt}{\textbf{Polynomial}} & LB & 1.23 & 1.32 & 1.32 & 12.5 & 16.2 & 34.5 & 48.7 \\ \cline{2-9}
\multicolumn{2}{|r|}{AFI (0.199s)} & 4 & 6 & 5 & 3 & 2 & 1 & 7 \\  
\multicolumn{2}{|r|}{PFI (5166s)} & 3 & 1 & 5 & 7 & 4 & 2 & 6 \\
 \hline
  \multirow{1}{4pt}{\textbf{RBF}} & LB & 69 & 71.5 & 73.7 & 74 & 75.4 & 76 & 76.7 \\ \cline{2-9}
\multicolumn{2}{|r|}{AFI (0.267s)} & 1 & 2 & 4 & 3 & 5 & 6 & 7 \\  
\multicolumn{2}{|r|}{PFI (11776s)} & 2 & 1 & 4 & 3 & 5 & 7 & 6 \\
 \hline
\end{tabular}
} \  \ \
\end{center}
\end{table}

We now compare our feature importance measure AFI with PFI (as implemented in \texttt{sklearn.inspection} with \texttt{n\_repeat} = 10). As a representative example, we show in Table~\ref{tbl:FI}, a comparison on SVMs trained on the Compas dataset. For each SVM model, in line `LB', we order the numerical features in the datasets (i.e., `priors\_count', `juv\_fel\_count', etc.) based on the verified individual fairness score with respect to the \textsc{NOISE} perturbation (with $\epsilon = 0.3$ in order to amplify the difference in score between the features). In lines `AFI' and `PFI' we show the order of these features based on the importance measured by AFI and PFI, respectively. We also indicate in parenthesis the time it took to compute these measures. We used the RAF+OH abstraction for AFI.
We can see that AFI better correlates with model variance to feature perturbations. In fact, the correlation is perfect in the linear SVM case. In the polynomial SVM case, the lower individual fairness scores indicate that the abstraction loses precision and this explains why AFI is less accurate (but still better than PFI, at least in identifying the least important feature). Note also that AFI is computed in a small fraction of time compared to PFI.
We show similar or better results for other datasets in Appendix~\ref{appendix-B}.

Finally, we recall that AFI can also measure the importance of tiers of a feature (cf. Section~\ref{sec:one-hot}). Thus, it offers another way to detect potential bias issues starting from discrepancies between tier importances (e.g., we observed that the tier `wife' of the `relationship' feature of the Adult dataset results has a different importance from the tier `husband').

\section{Conclusion}

We put forward a novel abstract feature importance measure based on an  abstract interpretation for SVMs tailored for achieving a precise symbolic representation of one-hot encoded features. We showed that our abstraction is effective for verifying robustness properties ---notably, individual fairness--- of SVMs and that our abstract feature importance measure outperforms the state-of-the-art.

As future work, we plan to extend our approach to verify alternative or stronger fairness notions. We also aim to design quantitative verification methods to provide probabilistic guarantees on the behavior of SVM models.

\subsubsection*{Acknowledgements}
Francesco Ranzato has been partially funded 
by the \emph{Italian MIUR}, under the PRIN2017 project no.\ 201784YSZ5 ``AnalysiS of PRogram Analyses (ASPRA)'', 
by \emph{Facebook Research}, under a 
``Probability and Programming Research Award'',
by an \emph{Amazon Research Award} for ``AWS Automated Reasoning'', and by a \emph{WhatsApp Research Award} for 
``Privacy aware program analysis''.

\clearpage
\appendix

\section{Appendix}
\label{appendix-A}

\subsection{Proofs}
\begin{proof}[Proof of Theorem~\ref{th:oh-alpha-prime}]
 We must show that $X = \gamma^{\OH_k}(a)$. Since we required values in $X$ to be the result of a one-hot encoding, values other than $f, t$ cannot occur. We start by calling $F_i = \{x_i | \vec{x} \in X \wedge x_i = f\}$ and $T_i = \{x_i | \vec{x} \in X \wedge x_i = t\}$ the sets of false and true values occurring the i-esim component, for every $1 \leq i \leq k$. We observe that $F_i$ can be either the singleton of $f$ or the empty set, the latter if and only if the i-esim member of the tier is never false in $X$. The same is true for $T_i$ being either the singleton of $t$ or the empty set.
 
 We now show that $X \subseteq \gamma^{\OH_k}(a)$, that is, every element of the former also belongs to the latter. If $X$ is the empty set condition is trivially satisfied. Otherwise, we consider a generic element $\vec{x'} \in X$. Since $\vec{x'}$ is the result of a one-hot encoding procedure, there exists $1 \leq i \leq k$ such that $x'_i = t$, while $x'_j = f$ for every $1 \leq j \leq k, j \neq i$. As a consequence, $T_i$ cannot be the empty set, and must be $T_i = \{t\}$, similarily we obtain $F_j = \{f\}$ for every $j \neq i$. This implies that $a_{i, t} = t$ and $a_{j, f} = f$, allowing to explicitly compute
 \begin{align*}
  \hat{\gamma}_i(a)
  =& \{\vec{x} \in \mathbb{R}^k | x_i \in \gamma^{\CP}(a_{i, t}) \wedge\\
   & \forall j \neq i\colon x_j \in \gamma^{\CP}(a_{j, f})\} \\
  =& \{\vec{x} \in \mathbb{R}^k | x_i \in \gamma^{\CP}(t) \wedge \forall j \neq i\colon x_j \in \gamma^{\CP}(f)\} \\
  =& \{\vec{x} \in \mathbb{R}^k | x_i \in \{t\} \wedge \forall j \neq i\colon x_j \in \{f\}\} \\
  =& \{\vec{x} \in \mathbb{R}^k | x_i = t \wedge \forall j \neq i\colon x_j = f\} \\
  =& \{\vec{x'}\}
 \end{align*}
 which in turns implies $\vec{x'} \in \hat{\gamma}_i(a) \subseteq \bigcup_{j = 1}^k \hat{\gamma}_j(a) = \gamma^{\OH_k}(a)$, hence $X \subseteq \gamma^{\OH_k}(a)$.
 
 Last, we show that $\gamma^{\OH_k}(a) \subseteq X$ by arguing that every element of the former also belongs to the latter. If $\gamma^{\OH_k}(a)$ is the empty set implication is trivially satisfied, otherwise we consider a generic element $\vec{x'} \in \gamma^{\OH_k}(a)$. Since $\gamma^{\OH_k}(a) = \bigcup_{i = 1}^k \hat{\gamma}_i(a)$, there must exists some $1 \leq z \leq k$ such that $\vec{x'} \in \hat{\gamma}_z(a)$, which can be written as
 \begin{align*}
  \hat{\gamma}_z(a)
  =& \{\vec{x} \in \mathbb{R}^k | x_z \in \gamma^{\CP}(a_{z, t}) \\
   & \wedge \forall j \neq z\colon x_j \in \gamma^{\CP}(a_{j, f})\} \\
  =& \{\vec{x} \in \mathbb{R}^k | x_z \in \gamma^{\CP}(\alpha^{\CP}(T_z)) \\
   & \wedge \forall j \neq z\colon x_j \in \gamma^{\CP}(\alpha^{\CP}(F_j))\} \\
 \end{align*}
 if $T_z = \varnothing$, or $\exists j \neq z\colon F_j = \varnothing$ then $\hat{\gamma}_z(a) = \varnothing$, which is not acceptable and thus must not be considered. As a consequence, it must be $T_z \neq \varnothing$ and $\forall j \neq z\colon F_j \neq \varnothing$, which in turns implies $T_z = \{t\}$ and $\forall j \neq z\colon F_j = \{f\}$. This allows to rewrite
 \begin{align*}
  \hat{\gamma}_z(a)
  =& \{\vec{x} \in \mathbb{R}^k | x_z \in \gamma^{\CP}(\alpha^{\CP}(T_z)) \\
   & \wedge \forall j \neq z\colon x_j \in \gamma^{\CP}(\alpha^{\CP}(F_j))\} \\
  =& \{\vec{x} \in \mathbb{R}^k | x_z \in \gamma^{\CP}(\alpha^{\CP}(\{t\})) \\
   & \wedge \forall j \neq z\colon x_j \in \gamma^{\CP}(\alpha^{\CP}(\{f\}))\} \\
  =& \{\vec{x} \in \mathbb{R}^k | x_z \in \gamma^{\CP}(t) \\
   & \wedge \forall j \neq z\colon x_j \in \gamma^{\CP}(f)\} \\
  =& \{\vec{x} \in \mathbb{R}^k | x_z \in \{t\} \wedge \forall j \neq z\colon x_j \in \{f\}\} \\
  =& \{\vec{x} \in \mathbb{R}^k | x_z = t \wedge \forall j \neq z\colon x_j = f\} \\
 \end{align*}
 which is a singleton. Since $\vec{x'} \in \hat{\gamma}_z(a)$, we can write $\hat{\gamma}_z(a) = \{\vec{x'}\}$. On the other hand, $T_z = \{t\}$ also implies $\exists \vec{x''} \in X\colon x''_z = t$. Moreover, it must be $x''_j = f$ for all $j \neq i$, as $\vec{x''}$ is the result of a one-hot encoding, hence $\vec{x''} = \vec{x'}$ and $\vec{x'} \in X$, which leads to the conclusion that every $\hat{\gamma}_i(a) \subseteq X$, hence $\gamma^{\OH_k}(a) \subseteq X$.
 
 Since both $X \subseteq \gamma^{\OH_k}(a)$ and $\gamma^{\OH_k}(a) \subseteq X$ hold at the same time, we can conclude that $X = \gamma^{\OH_k}(a)$ and $a$ precisely $X$ without loss of precision.
\end{proof}

\begin{proof}[Proof of Theorem \ref{th:oh-f-sound}]
 It must be shown that, for any $f: \mathbb{R} \rightarrow \mathbb{R}$ and for any $a \in \OH_k$, $f(\gamma^{\OH_k}(a)) \subseteq \gamma^{\OH_k}(f^{\OH}(a))$.
 \begin{align*}
   & f(\gamma^{\OH_k}(a)) = f \left( \bigcup_{i = 1}^k \hat{\gamma}_i(a) \right) \\
  =& \bigcup_{i = 1}^k f(\hat{\gamma}_i(a)) \\
  =& \bigcup_{i = 1}^k \{ f(\vec{x}) | x_i \in \gamma^{\CP}(a_{i, t}) \wedge \\
   & \forall j \neq i\colon x_j \in \gamma^{\CP}(a_{j, f}) \} \\
  =& \bigcup_{i = 1}^k \{ \vec{x} | x_i \in f(\gamma^{\CP}(a_{i, t})) \wedge \\
   & \forall j \neq i\colon x_j \in f(\gamma^{\CP}(a_{j, f})) \} \\
  \subseteq& \bigcup_{i = 1}^k \{ \vec{x} | x_i \in \gamma^{\CP}(f^{\CP}(a_{i, t})) \wedge \\
   & \forall j \neq i\colon x_j \in \gamma^{\CP}(f^{\CP}(a_{j, f})) \} \\
  =& \bigcup_{i = 1}^k \hat{\gamma}_i(f^{\OH}(a)) = \gamma^{\OH_k}(f^{\OH}(a))
 \end{align*}
\end{proof}

\begin{proof}[Proof of Corollary \ref{th:oh-f-complete}]
(i) By assuming no $\top^{\CP}$ value occurs in $a$, then $f(\gamma^{BC}(a_{i, z})) = \gamma^{BC}(f^{BC}(a_{i, z}))$ for any $1 \leq i \leq k$ and $z \in \{f, t\}$. This equivalence can be exploited in proof of Th. \ref{th:oh-f-sound}, yielding $f(\gamma^{\OH_k}(a)) = \gamma^{\OH_k}(f^{\OH}(a))$.

(ii)
 We will prove the stronger property that analysis is complete and no $\top^{\CP}$ values can be generated. Proof is by induction on the number $n \in \mathbb{N}$ of applications of functions $f_i$.
 \begin{description}
  \item[$P(0)$] Due to Th.\ref{th:oh-alpha-prime}, it is possible to abstract $X$ with some $a_0 \in \OH_k$ without loss of precision. Remark \ref{th:oh-alpha-no-top} also applies and guarantees that no $\top^{\CP}$ can occur in $a_0$.
  \item[$P(n) \Rightarrow P(n + 1)$] By inductive hypotheses we have an intermediate abstract value $a_n \in \OH_k$ which represents its concrete counterpart without loss of precision, and we also know that $a_n$ does not contain any $\top^{\CP}$ values. We can therefore compute $a_{n + 1} = f_{n + 1}^{\OH_k}(a_n)$ and, due to Corollary \ref{th:oh-f-complete}, no loss of precision can occur in $a_{n +1}$. Also, by definition of $f^{\OH}$ in Th.\ref{th:oh-f-sound}, $\top^{\CP}$ cannot be introduced unless already present in $a_n$, which is not the case thanks to inductive hypotheses.
 \end{description}
 Alternatively, it is possible to consider $f = f_1 \circ f_2 \circ \ldots \circ f_n$ and directly apply Corollary \ref{th:oh-f-complete}, although this appears more restrictive as it does not allow for composition with other complete transfer functions in between.
\end{proof}

\subsection{Additional Experimental Data}
\label{appendix-B}

\noindent
\begin{table*}
\begin{center}
  \captionof{table}{Accuracy, balanced accuracy, and individual fairness scores for SVMs trained on each dataset.\label{Tab:OH}}
\begin{tabular}{ |p{1cm}|p{1.7cm}|p{0.5cm}|p{0.5cm}|p{0.5cm}|p{0.5cm}|p{0.5cm}|p{0.5cm}|p{0.5cm}|p{0.5cm}|p{0.5cm}|p{0.5cm}|p{0.5cm}|p{0.5cm}|p{0.5cm}|p{0.5cm}|  }
 \hline
 \multirow{2}{4pt}{\textbf{Dataset}} & \multirow{2}{4pt}{\textbf{Kernel}} & \multirow{2}{4pt}{\textbf{Acc.}} & \multirow{2}{4pt}{\textbf{Bal. Acc}} &  \multicolumn{3}{c|}{\textbf{Interval}} & \multicolumn{3}{c|}{\textbf{Interval OH}} & \multicolumn{3}{c|}{\textbf{RAF}} & \multicolumn{3}{c|}{\textbf{RAF OH}} \\ \cline{5-16}
 & & & & C & NC & N & C & NC & N & C & NC & N & C & NC & N\\
 \hline\hline
   \multirow{3}{4pt}{Adult} & L(1)& 84.6 & 75.6 & 95.2 & 91.6 & 96.5 & 95.2 & 91.6 & 96.5 & 95.2 & 91.6 & 96.5 & 95.2 & 91.6 & 96.5 \\
 &R(0.05,0.01) & 83.8 & 72.0 & 2.8 & 0.0 & 2.4 & 2.8 & 0.0 & 2.4 & 42.4 & 37.2 & 94.8 & 97.9 & 95.4 & 97.5 \\
 &P(0.01,3,3) & 83.9 & 76.7 & 0.0 & 0.0 & 0.0 & 0.0 & 0.0 & 0.0  & 0.5 & .03 & 0.5 & 0.5 & 0.03 & 0.5 \\
 \hline
 \multirow{3}{4pt}{Compas} & L(1)& 64.7 & 64.1 & 99.5 & 94.9 & 95.5 & 99.5 & 94.9 & 95.5 & 99.5 & 94.9 & 95.5 & 99.5 & 94.9 & 95.5 \\
 &R(0.05,0.01) & 64.5 & 63.1 & 42.5 & 1.0 & 54.3 & 42.5 & 1.0 & 54.3 & 71.6 & 66.9 & 91.8 & 97.5 & 89.3 & 94.4 \\
 &P(0.01,3,3) &64.3 & 63.9 & 0.0 & 0.0 & 0.0 & 0.0 & 0.0 & 0.0 & 0.6 & 0.1 & 0.5 & 0.6 & 0.1 & 0.5  \\
 \hline
 \multirow{3}{4pt}{Crime} & L(1)& 82.0 & 82.0 & 0.75 & 0.25 & 67.2 & 60.9 & 29.6 & 67.2 & 0.75 & 0.25 & 67.2 & 60.9 & 29.6 & 67.2\\
 &R(1,$10^{-3}$) &77.7 & 77.7 & 1.0 & 0.0 & 37.8 & 92.5 & 29.3 & 37.8 & 22.1 & 16.0 & 90.2 & 100 & 91.0 & 90.2\\
 &P(1,9,0) &74.2 & 74.1 & 71.9 & 9.0 & 13.3 & 71.9 & 9.0 & 13.3 & 83.0 & 31.3 & 55.4 & 93.5 & 39.1 & 55.4\\
 \hline
   \multirow{3}{4pt}{German} & L(1)& 79.0 & 70.8 & 94.5 &  81.5 & 87.5 & 94.5 & 81.5 & 87.5  & 94.5 & 81.5 & 87.5 & 94.5 & 81.5 & 87.5 \\
 &R(10,0.05) &79.5 & 74.1 & 0.0 & 0.0 & 0.0 & 0.0 & 0.0 & 0.0 & 0.0 & 0.0 & 2.0 & 82.0 & 0.0 & 2.0 \\
 &P(0.01,6,6) &75.5 & 71.8 & 0.0 & 0.0 & 0.0 & 0.0 & 0.0 & 0.0  & 76.0 & 10.0 & 78.5 & 76.0 & 10.0 & 78.5 \\
 \hline
  \multirow{3}{4pt}{Health} & L(0.01)& 77.8 & 69.5 & 91.3 & 90.6 & 99.4 & 96.1 & 95.5 & 99.4 & 91.3 & 90.6 & 99.4 & 96.1 & 95.5 & 99.4\\
 &R(0.1,0.01) &82.3 & 76.4 & 0.68 & 0.65 & 1.26 & 3.38 & 2.87 & 4.71 & 0.82 & 0.81 & 80.1 & 94.1 & 29.4 & 82.0 \\
 &P(0.1,3,0.01) &71.0 & 61.0 & 0.0 & 0.0 & .005 & 0.0 & 0.0 & .005 & 0.03 & 0.02 & 6.62 & 0.03 & 0.02 & 6.62\\
 \hline
\end{tabular}
\end{center}
\end{table*}

Table \ref{Tab:OH} shows accuracy, balanced accuracy, and individual fairness scores for SVMs with linear kernels (represented as L(reg. param.)), polynomial kernels (represented as P(reg. param, degree, base)), and RBF kernels (represented as R(reg. param, $\gamma$)). Individual fairness scores are computed with respect to the \textsc{NOISE} (N), \textsc{CAT} (C), and \textsc{NOISE-CAT} (NC) similarity relations. The table also compares results with and without the $\OH$ abstraction.

\begin{table*}[t]
\caption{Comparison of our AFI and PFI on Adult.} \label{tab:Adult}
\begin{center}
\begin{tabular}{ |p{3cm}|p{0.75cm}|p{0.75cm}|p{0.75cm}|p{0.75cm}|p{0.75cm}|p{0.75cm}|p{0.75cm}|}
 \hline
\multirow{1}{4pt}{\textbf{Linear}~($\epsilon~=~0.3$)} & LB & 83.4 & 96 & 98.37 & 98.37 & 98.44 & 99.2 \\ \cline{2-8}
\multicolumn{2}{|r|}{AFI} & 1 & 2 & 3 & 4 & 5 & 6 \\  
\multicolumn{2}{|r|}{PFI} & 1 & 2 & 6 & 4 & 5 & 3 \\
 \hline
 \multirow{1}{4pt}{\textbf{Polynomial}~($\epsilon~=~0.1$)} & LB & 23.2 & 23.9 & 24.5 & 28.3 & 29.5 & 95.2 \\ \cline{2-8}
\multicolumn{2}{|r|}{AFI} & 2 & 5 & 6 & 3 & 4 & 1  \\  
\multicolumn{2}{|r|}{PFI} & 6 & 5 & 4 & 2 & 3 & 1  \\
 \hline
  \multirow{1}{4pt}{\textbf{RBF}~($\epsilon~=~0.3$)} & LB & 50.3 & 53.5 & 56.4 & 56.6 & 57.7 & 57.9  \\ \cline{2-8}
\multicolumn{2}{|r|}{AFI} & 1 & 2 & 3 & 4 & 5 & 6  \\  
\multicolumn{2}{|r|}{PFI} & 2 & 1 & 4 & 5 & 6 & 3  \\
 \hline
\end{tabular}
\end{center}
\end{table*}

\begin{table*}[t]
\caption{Comparison of our AFI and PFI on German.} \label{tab:German}
\begin{center}
\begin{tabular}{ |p{3cm}|p{0.5cm}|p{0.5cm}|p{0.5cm}|p{0.5cm}|p{0.5cm}|p{0.5cm}|p{0.5cm}|p{0.5cm}|p{0.5cm}|p{0.5cm}|}
 \hline
\multirow{1}{4pt}{\textbf{Linear}~($\epsilon~=~0.3$)} & LB & 78 & 83.5 & 87 & 89.5 & 90 & 90 & 98.5 & 98.5 & 99.5 \\ \cline{2-11}
\multicolumn{2}{|r|}{AFI} & 1 & 2 & 3 & 4 & 5 & 6 & 7 & 8 & 9 \\  
\multicolumn{2}{|r|}{PFI} & 3 & 7 & 1 & 6 & 4 & 2 & 8 & 5 & 9 \\
 \hline
 \multirow{1}{4pt}{\textbf{Polynomial}~($\epsilon~=~0.3$)} & LB & 83 & 86 & 86.5 & 87 & 92.5 & 92.5 & 93 & 94.5 & 94.5 \\ \cline{2-11}
\multicolumn{2}{|r|}{AFI} & 1 & 6 & 5 & 3 & 2 & 9 & 4 & 7 & 8 \\  
\multicolumn{2}{|r|}{PFI} & 2 & 3 & 4 & 5 & 9 & 1 & 6 & 7 & 8 \\
 \hline
  \multirow{1}{4pt}{\textbf{RBF}~($\epsilon~=~0.1$)} & LB & 54 & 54 & 54.5 & 55 & 56.5 & 57 & 57.5 & 58 & 58.5 \\ \cline{2-11}
\multicolumn{2}{|r|}{AFI} & 1 & 2 & 3 & 4 & 5 & 6 & 8 & 7 & 9 \\  
\multicolumn{2}{|r|}{PFI} & 1 & 3 & 7 & 9 & 4 & 5 & 8 & 2 & 6 \\
 \hline
\end{tabular}
\end{center}
\end{table*}

We show further comparisons between our proposed AFI and PFI in Table~\ref{tab:Adult} and Table~\ref{tab:German}. We compute individual fairness scores with respect to the \textsc{NOISE} perturbation. We choose different $\epsilon$ values to amplify the difference in score between the features. Note the much better correlation of AFI with fairness scores than PFI. We do not report results for the Crime dataset as it has dozens of numerical features. PFI timed out after executing for several hours on Health.

\end{document}